\documentclass[review]{elsarticle}

\usepackage{lineno,hyperref}
\modulolinenumbers[5]
 
\journal{Pattern Recognition}
 
\newcommand{\etal}{\mbox{\textit{et al. }}}
 
\usepackage[cmex10]{amsmath}
\usepackage{epstopdf}
\usepackage{algorithmic}
\usepackage[ruled,vlined]{algorithm2e}
\usepackage{url}
\usepackage[dvipsnames]{xcolor}
\usepackage{multirow}
\usepackage{hhline}
\usepackage[shortlabels]{enumitem}
\usepackage{pbox}
\usepackage{rotating}
\usepackage[export]{adjustbox}
\usepackage[skip=0pt]{caption}
\usepackage{booktabs}
\usepackage{amsmath}
\usepackage{amssymb}
\usepackage{subfigure}
 
\biboptions{sort&compress}
 
\begin{document}
 
\begin{frontmatter}

\title{Human Interaction Recognition Framework based on Interacting Body Part Attention}
 
%
\author[add1]{Dong-Gyu Lee}
\author[add2]{Seong-Whan Lee\corref{corauth}}
\cortext[corauth]{Corresponding author}
\ead{sw.lee@korea.ac.kr}
 
\address[add1]{Department of Artificial Intelligence, Kyungpook National University,\\ Daehak-ro, Buk-gu, Daegu 41566, Korea}
\address[add2]{Department of Artificial Intelligence, Korea University,\\ Anam-ro, Seongbuk-gu, Seoul 02841, Korea}

\begin{abstract}
Human activity recognition in videos has been widely studied and has recently gained significant advances with deep learning approaches; however, it remains a challenging task. In this paper, we propose a novel framework that simultaneously considers both implicit and explicit representations of human interactions by fusing information of local image where the interaction actively occurred, primitive motion with the posture of individual subject's body parts, and the co-occurrence of overall appearance change.
Human interactions change, depending on how the body parts of each human interact with the other. The proposed method captures the subtle difference between different interactions using interacting body part attention. Semantically important body parts that interact with other objects are given more weight during feature representation. The combined feature of interacting body part attention-based individual representation and the co-occurrence descriptor of the full-body appearance change is fed into long short-term memory to model the temporal dynamics over time in a single framework. We validate the effectiveness of the proposed method using four widely used public datasets by outperforming the competing state-of-the-art method.
\end{abstract}
 
 
\end{frontmatter}
 
 
\section{Introduction}
\label{Intro}
Despite many efforts in the last decades, understanding human activity from a video remains a developing and challenging task in the field of computer vision.
In particular, human interaction recognition is a key component of video understanding and is indispensable because it is frequently observed in the real video. It has various applications, such as video surveillance systems, a human-computer interface, automated driving, an understanding of human behavior, and prediction.
 
The primary reason for the difficulty in human interaction recognition is that we need to consider both single actions and co-occurring individual activities of people to understand the complex relationship between the participants. The key to the success of this task is how to extract discriminative features that can effectively capture the motion characteristics of each person in space and time.
The general approach is the implicit representation of the video, such as a bag-of-words (BoW)-based model \cite{laptev2008learning}. The BoW approach describes the entire frame by clustering spatio-temporal image patch features, which are extracted from interest points \cite{laptev2005space}, preset attributes, or key-poses from videos \cite{ryoo2011human, kong2016max, ziaeefard2015semantic}. In recent studies, deep neural network-based video classification methods, such as 3D convolutional neural networks (CNN) \cite{ji20133d}, two-stream CNN \cite{simonyan2014two}, or multi-stream CNN \cite{tu2018multi}, have shown promising results for video representation. One advantage of this type of approach is that the representation is robust to the failure of key-point extraction because it is the overall distribution of pixels that compose the entire image, not a specific point. 
However, this approach lacks high-level information, which can be a crucial property of human behavior understanding. The implicit representation of the overall frame is relatively overlooked, compared to the underlying explicit properties of the interaction, such as ``where/when the feature was extracted'' or ``who moved how''.
 
In contrast to single action moves only or group activities focusing on relationships between multiple objects, human interaction includes both individual motions and relationships, and both are equally important. The acting objects directly influence the interacting objects and the reaction is decided, depending on the acting object's individual motion.
In a case of interaction, subtle differences, such as, how the body parts of each person interact with others' body parts, can change the activity class.
 
\begin{figure}
\centering
\subfigure[Handshake]{
\includegraphics[height=0.25\linewidth]{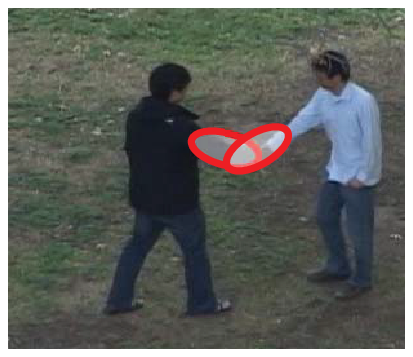}\label{Handshake}}
\subfigure[Punch]{
\includegraphics[height=0.25\linewidth]{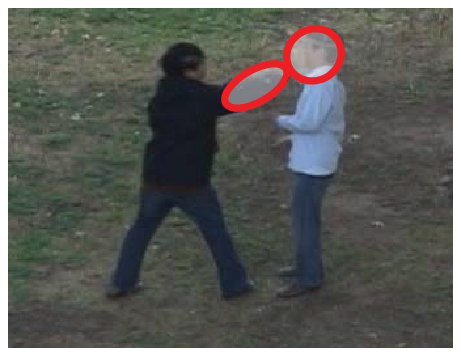}\label{Punch}}
\subfigure[Hug]{
\includegraphics[height=0.25\linewidth]{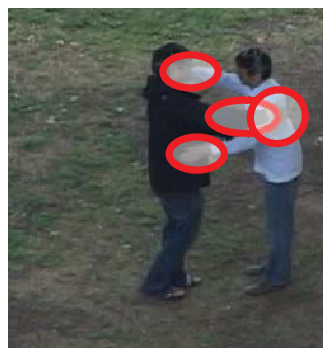}\label{Hug}}
\caption{An example of the subtle differences in three activities: the right hand of the person on the left stretching to three different body parts of the other person.}
\label{Intro}
\end{figure}
 
Fig. \ref{Intro} shows an example of three different human interactions that include the ``stretch hand'' motion. In the three pictures, the person on the left stretches his right hand forward; however, the type of interaction varies, depending on which body part of the person on the right interacts with it. In the case of a handshake in \ref{Handshake}, the right hand of the person on the left interacts with the right hand of the person on the right; however, a punch in \ref{Punch} can be understood as an interaction between a hand and the head of two different persons, at a faster speed. Although the hug includes the stretching motions of the right hand, the left hand and the body also move forward in \ref{Hug}. Thus, it can be understood as an interaction between the torsos and the hands of two person. As shown above, the type of human interaction can be distinguished by the subtle difference in these detailed movements. The attribute factor such as ``which movements are meaningful?'' or ``how do they react with each other?'' should be considered. 
However, the important attribute of the interaction is a problem-dependent factor; thus, automatic detection of these attributes, with a high-level of understanding, is necessary for efficient and robust human interaction recognition.

To address this problem, we exploit a body joint extraction method to specify the interacting parts. 
The explicit representation of human behavior provides a clearer interpretation of primitive human body pose and movement.
Because not all joints are informative for human behavior, some less-relevant movement of joints may generate noise which can decrease clarity of behavior representation. Thus, the proposed method is designed to detect the informative part of the interaction. 
The joint-based feature and local image patch feature in the interacting joint are weighted in proportion to the activation degree of the interacting body part.
The lack of high-level information from implicit representation supplemented by using the extracted body joints.

Human behavior can be represented by a combination of movements of body joints in 3D space. The release of cost-effective RGB-D sensors has encouraged plenty of skeleton based human activity recognition methods. The dynamic human skeletons usually convey significant information that is complementary to others \cite{liu2019rgb, song2017end, shi2019skeleton, liu2018skeleton, liu2018structured, liu2020hds}.
In recent studies, the deep learning-based human pose estimation at RGB video has achieved excellent advances \cite{sun2019deep, wei2016convolutional, cao2018openpose, toshev2014deeppose}. 
Thus, it has become possible to utilize the advantage of dynamic human skeleton information while using widely used RGB sensors in real-time. The exploitation of an estimated skeleton can provide sophisticated information about articulated human poses with fewer environmental constraints.
Although the pose estimation result from the RGB sensor can provide significant information about human behavior, it is still rougher information than the RGB-D sensor-based skeleton estimation. The partial missing of the joints estimation often occur due to camera angle or occlusion. It reduces the quality of skeleton-based human behavior modeling. We designed the body joint extraction procedure from the pose estimation result to exploit the imperfect body joint information for the rest parts of the proposed framework.

In this paper, we present a novel human activity recognition framework that can consider the interacting body parts with local image patches based on the imperfect state joints, and the full-body image, simultaneously. The interacting body part attention allows the proposed framework to selectively focus on informative points of interaction in each video, automatically. Focusing on important joints reduces the influence of the less important joints on the descriptor. The spatial ambiguity of the local image patch is minimized by combining local image feature with body joint information while retaining the advantage of implicit representation. The full-body image represents the co-occurrence of individual human action with the importance of the sub-volume at each time step. The movement of five body parts is explicitly represented with motion and posture feature representation.

In summary, our three main contributions to this paper are as follows: (1) We develop an interacting body part attention mechanism, which can selectively focus on the important local movement of the participants. The body joint, in its imperfect state, can be exploited in this method. (2) We propose a framework that simultaneously considers both implicit and explicit representation of human interaction by fusing local image patch features where the interactions actively occurred, local motion with posture feature, and co-occurrence of overall appearance of multi-person. Explicit and implicit representations work complementarily to obtain a fine-grained representation of human interaction. (3) We demonstrate that the proposed method achieves state-of-the-art performance and shows its extensibility.
The subsequent sections of the paper are organized as follows: In Section \ref{Related}, we present the studies related to our work. We discuss the overall model in Section \ref{Proposed}. The experimental results are detailed in Section \ref{Experiment}. Finally, we present the conclusions of this work in Section \ref{conclusion}.
 
\section{Related Works}
\label{Related}
In this section, we briefly review some of the literature that is relevant to our work.
\subsection{Human activity recognition from RGB video}
The recognition in RGB video is the most representative and traditional study and has a wide range of applications \cite{ziaeefard2015semantic, liu2019rgb, bulthoff2003biologically, xi2002facial, kang2014nighttime, park2013face, roh2010view, kim2018discriminative}.
Conventional methods for human activity recognition typically use hand-crafted features to represent the video. Many studies emphasized the importance of spatio-temporal local features. The BoW-based paradigm, with 3D XYT volume, is one of the most popular approaches \cite{laptev2008learning, laptev2005space, ryoo2011human, kong2016max, ziaeefard2015semantic}. The key attributes of interacting people were captured using preset motion attributes, such as interactive phrases \cite{kong2014interactive}, discriminative key components \cite{sefidgar2015discriminative}, action attributes \cite{liu2011recognizing}, or poselets \cite{raptis2013poselet}. Kong \etal \cite{kong2014interactive} proposed multiple interactive phrases as the latent mid-level feature to represent human interaction from individual actions. Sefidgar \etal \cite{sefidgar2015discriminative} represented the interaction by a set of key temporal moments and the spatial structures they entail. Despite these studies, human interaction has a subtle difference in detailed movements, such as between a punch and a pat, which is too small to be described by a specific number of phrases. 
 
For the past few years, the deep learning-based video representation methods, such as 3D CNN \cite{ji20133d}, two-stream CNN \cite{simonyan2014two}, and multi-stream CNN \cite{tu2018multi} have shown their effectiveness in video representation by overcoming the problem-dependent limitation of the hand-crafted features. Ibrahim \etal \cite{ibrahim2016hierarchical} captured the temporal dynamics of the whole activity based on the dynamics of the individual people. A combination of multi-level representations is modeled with a temporal link. 
Shu \etal \cite{shu2019hierarchical} proposed a hierarchical long short-term concurrent memory (H-LSTCM) to effectively address the problem of human interaction recognition with multiple persons by learning the dynamic inter-related representation over time. They aggregated the inter-related memory from individuals by capturing the concurrently long-term inter-related dynamics among multiple persons rather than the dynamics of individuals.
Tang \etal \cite{tang2019coherence} proposed coherence constrained graph LSTM (CCG-LSTM) to learn the discriminative representation of a group activity. The CCG-LSTM modeled the motion of individuals relevant to the whole activity, while suppressing the irrelevant motions.
Shu \etal \cite{shu2017concurrence} proposed concurrence-aware long short-term sub-memories to model the long-term inter-related dynamics between two interacting people. Lee \etal \cite{lee2019prediction} proposed a human activity representation method, based on the co-occurrence of individual action, to predict partially observed interaction with the aid of a pre-trained CNN. Mahmood \etal \cite{mahmood2018robust} segmented full-body silhouettes and identified key body points to extract features that had distinct characteristics. Deng \etal \cite{deng2016structure} utilized CNN to obtain a single person action label and a group activity label and explore the relation between the actions of all individuals. However, because most studies focused on modeling individuals of the relationships in the scene, attention to subtle movements of individual human behavior and the other person's detailed reactions was relatively less considered.
 
\subsection{Human activity recognition using RGB-D sensor}
With the advance of RGB-D sensors, such as Microsoft Kinect, Asus Xtion, and Intel RealSense, action recognition using 3D skeleton sequences has received significant attention and many advanced approaches have been proposed during the past few years \cite{liu2019rgb, liu2018structured,liu2020hds, liu2018skeleton, song2017end, shi2019skeleton}. 
Liu \etal \cite{liu2018structured} proposed a novel kernel enhanced bag of semantic moving words (BSW) approach to represent the dynamic property of skeleton trajectories. A structured multi-feature representation is composited by aggregating BSW with the geometric feature of skeleton data.
In other work, a novel HDS-SP descriptor that consists of both spatial and temporal information from specific viewpoints was proposed by Liu \etal \cite{liu2020hds}. The HDS-SP projects 3D trajectories on specific planes and creates reasonable histograms. The most suitable viewpoint is sought out through both local search and particle swarm optimization.
In other recent studies of skeleton-based human action recognition, attention-based human action recognition was attempted, owing to greater accessibility of individual body parts. Liu \etal \cite{liu2018skeleton} proposed a new class of long short-term memory (LSTM) \cite{hochreiter1997long}, which is capable of selectively focusing on the informative joints in each frame, using a global context memory cell. Song \etal \cite{song2017end} proposed an end-to-end framework with two types of attention modules, a spatial attention module and temporal attention module, which allocate different attention to different joints and frames, respectively.
 
Recently, a graph neural network (GCN)-based approach received significant attention for skeleton representation, owing to the kinematic dependency between the joints of the human body. Shi \etal \cite{shi2019skeleton} proposed a directed graph neural network to extract the dependencies of joints, bones, and their relationships, for the final action recognition task. Yang \etal \cite{yang2020feedback} achieved outstanding performance, based on a feedback graph convolutional network, which adopted a feedback mechanism to the GCN. Multi-stage temporal sampling and a dense connections-based feedback graph convolutional block is proposed to introduce feedback connections into the GCN.
Although a human skeleton can provide sophisticated information about human behavior, to maximize the effectiveness of a skeleton-based modeling method, accurate and delicate skeleton joint information is essential. However, most of cost efficient RGB-D sensors are currently limited to indoor applications in close distance. The imperfect joint state, such as missing joints or jittering joints, is often observed when the joints are estimated from RGB video in the wild.
 
\subsection{Pose estimation}
With the success of the new era of deep learning-based approaches, some stuides have achieved excellent results in extracting the human body joints from RGB videos through pose estimation \cite{sun2019deep, wei2016convolutional, yang2007reconstruction, cao2018openpose, toshev2014deeppose, roh2007accurate}. Toshev \etal \cite{toshev2014deeppose} proposed the first deep neural network-based pose estimation method, called DeepPose. They formulated the pose estimation problem to a CNN-based regression problem toward body joints. Wei \etal \cite{wei2016convolutional} proposed a convolutional pose machine (CPM), which consists of an image feature computation module, followed by a prediction module. The CPM can handle long-range dependencies between variables in a structured prediction task by designing a sequential architecture, composed of convolutional networks that directly operate on a belief map from the previous stage. Sun \etal \cite{sun2019deep} proposed the HRNet model, which has achieved great performance by maintaining a high-resolution representation throughout the process. Cao \etal \cite{cao2018openpose} successfully estimated body joints using part affinity fields (PAFs). The proposed PAF can efficiently detect the 2D pose of multiple people in real-time with high accuracy, regardless of the number of people, using the bottom-up approach. The proposed method uses a nonparametric representation to learn to associate body parts with individuals in the image. In this paper, we utilized OpenPose method \cite{cao2018openpose} to estimate initial human body joint from video, to take advantage of its high accuracy and real-time performance, which are essential properties for practical real-world applications.
 
Because pose estimation and action recognition are two closely related problems, there are some studies that have performed both tasks simultaneously. Luvizon \etal \cite{luvizon20182d} proposed a multi-task deep learning approach and performed joint 2D and 3D pose estimation from still images and human action recognition in a single framework. Nie \etal \cite{xiaohan2015joint} proposed an AND-OR graph-based action recognition approach, which exploited the hierarchical part composition analysis. Even though the end-to-end approach has advantages regarding task optimization, it has limited extensibility to videos in varying real-world environments, such as activity scenarios. Furthermore, an approach to research involving interactions, rather than single human actions, is methodologically distinct. 
 
 
\section{Proposed Method}
\label{Proposed}
In this section, we present the details of the proposed framework for human interaction recognition based on interacting body part attention. The overview of framework is shown in Fig. 2.
From a video, we extract human body joints based on combination of object detection and pose estimation. We then pass it through three different streams: full-body feature stream, local image patch feature stream, and motion and posture feature stream.
The motion feature and the local image patch feature were concatenated and they feed into the sub-LSTM after it is weighted by interacting body part attention.
The outputs of the sub-LSTM and the co-occurrence descriptor are concatenated, then fed into the LSTM for final classification.
 
\begin{figure}
\centering
\includegraphics[width=1.0\linewidth]{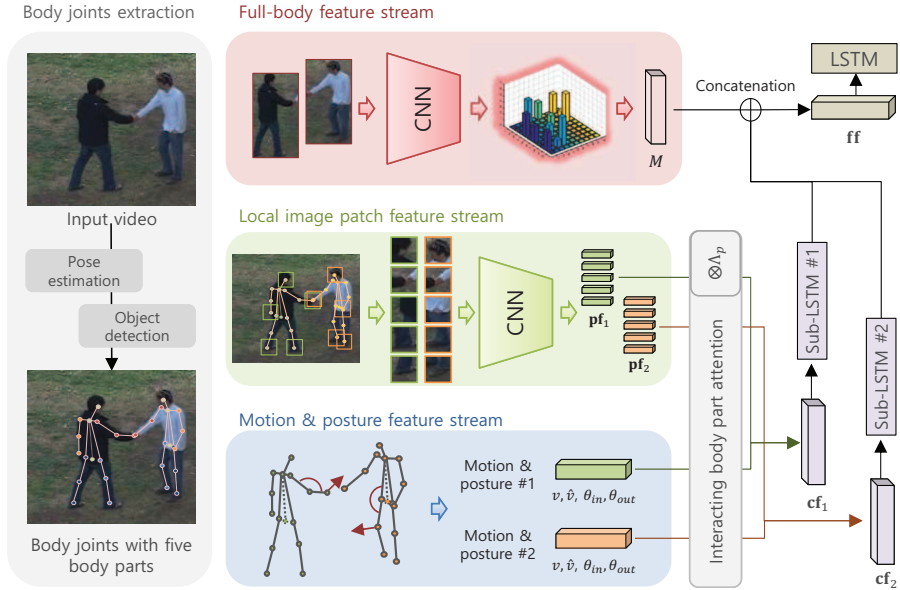}
\caption{Overview of the proposed framework. The extracted body joints with five body parts are utilized in each different stream for the interacting body part attention.}
\label{framework}
\end{figure}
 
\subsection{Body joint extraction}
From a given video clip, to estimate a multi-person body joint, we first utilize a publicly available pre-trained object detection model to obtain the bounding box of the object; we used the Faster-RCNN \cite{ren2015faster} with the Inception-resnet-v2 network \cite{szegedy2017inception}. The detection result provides $(x, y)$ coordinates with the height and width of each object. We also perform initial pose estimation using a publicly available real-time multi-person 2D pose estimation method, based on bottom-up approach, i.e., OpenPose \cite{cao2018openpose}. OpenPose provides the first combined body and foot key-point detector, which enables us to obtain joints even though the links between joints are disconnected.
 
\begin{figure}
\centering
\subfigure[Index of initial pose estimation results]{
\includegraphics[height=0.5\linewidth]{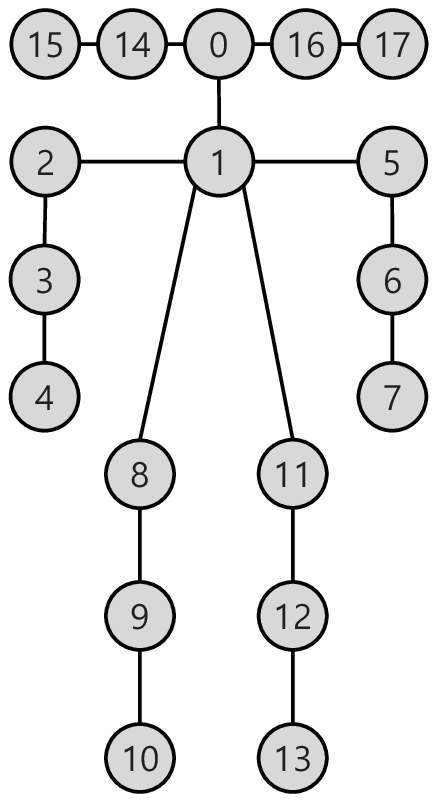}\label{Estipose}}
\hspace{1em}
\subfigure[Index of body joints with five body parts]{
\includegraphics[height=0.5\linewidth]{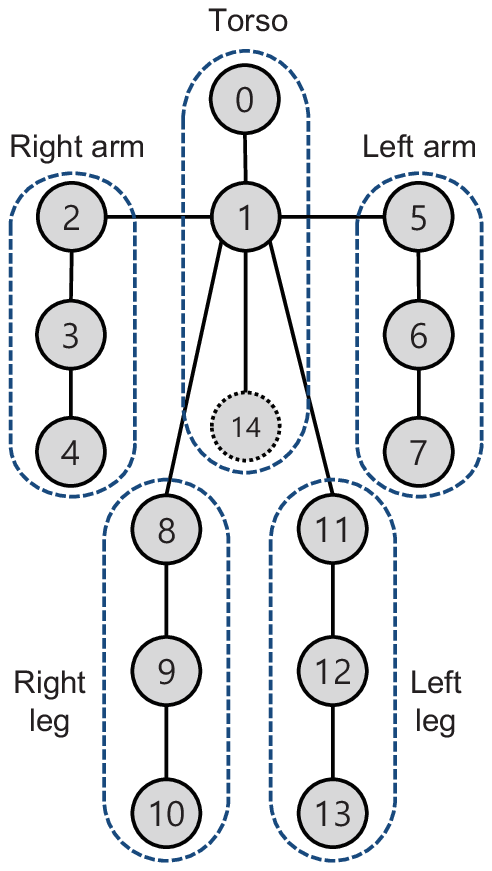}\label{bodyparts}}
\caption{Illustration of the index of initial pose estimation results and the configuration of the human body pose, divided into five body parts (torso, left arm, right arm, left leg, and right leg).}
\label{bodypartGenerates}
\end{figure}

The initial pose estimation result consists of eighteen joints for each human object, from $0$ to $17$. The composition of the joints of the estimated pose is shown in Fig. \ref{Estipose}. 
We first select fifteen joints, out of nineteen, to generate five body parts.
For each human subject, we convert each joint $j^{ori}_{i} = \{j^{ori}_{0}, ... , j^{ori}_{17}\}$ to $j_{i} = \{j_{0}, ... , j_{14}\}$, as illustrated in Fig. \ref{bodyparts}.
Additionally, the coordinate of the center of the joints between $j^{ori}_8$ and $j^{ori}_{11}$ is designated as joint $j_{14}$, for utilization of the torso information; this is referred to as the hip.
The five body parts consist of three different joints, from $j_0$ to $j_{14}$, defined as follows: right arm $p_1$=($j_2$, $j_3$, $j_4$), left arm $p_2=$($j_5$, $j_6$, $j_7$), right leg $p_3=$($j_8$, $j_9$, $j_{10}$), left leg $p_4=$($j_{11}$, $j_{12}$, $j_{13}$), and torso $p_5=$($j_0$, $j_1$, $j_{14}$). Each number denotes the joint index and their 2D coordinates are defined as $p_{pi,pj,x,y}$, where $pi$ denotes the body part index and $pj$ denotes the joint index in each body part. 
 
Although the pose estimation shows promising results, it often fails to detect body joints still at an individual frame-level. We performed two additional procedures to increase the usability and dependability of body joints to manage the noisy pose estimation.
To complement the occasional absence of joint information, if the previous $l$ frames have failed to estimate a joint, we restore the missing joint by linearly interpolating the coordinate between the current joint at $t$ and the last successfully detected joint at the $t-l$ frame. 
We also utilized the bounding box from the human detection to filter out some residual, or unsuccessful, body joint estimation results, the majority of which are the partial detection of a human object, due to occlusion. First, both the head and torso of each object must be included in the bounding box. If an estimation failure occurs about the head position $j^{ori}_0$, the average value of the non-failure joint among \{$j^{ori}_{14}, j^{ori}_{15}, j^{ori}_{16}, j^{ori}_{17}$\} is used as the head position, $j_0$. By performing two additional post-processing steps, we obtained the stability and reliability of the body joints from real-time multi-person pose estimation results.

\subsection{Body joint-based local features}
{\bf{Motion and posture feature:}}
To express the overall motion and posture of an individual human behavior, we extract two types of motion features and two types of posture features based on body part status.
At each time step, for fifteen body joints of five body parts, the average value of the 2D coordinate difference between each previous point and each current point of the sequence is used to calculate the motion velocity $v_p^t$ and acceleration $\hat{v}_p^t$.
This is a simple, but effective, way to express the explicit motion of the body parts and joints by changing the position of each joint.
We also calculate the inner angle within the body parts to represent the relative position of the joint inside the body part. The angle of each part $\theta_{in}$ is calculated using the following equation:
\begin{equation}
\label{angle}
\begin{aligned}
{\bf{a}} = (p_{pi,1,x} - p_{pi,2,x}, p_{pi,1,y} - p_{pi,2,y}), \\
{\bf{b}} = (p_{pi,1,x} - p_{pi,3,x}, p_{pi,1,y} - p_{pi,3,y}), \\
\theta=arccos\left(\frac{\bf{a}\cdot\bf{b}}{|\bf{a}||\bf{b}|}\right). \\
\end{aligned}
\end{equation}
 
We also calculate the angle between body parts using Eq. (\ref{angle}). The outer angle $\theta_{out}$ denotes the angle between the connected body part, which is calculated using the following joint indexes, as the input in each frame: $p^{out}_1$=($j_1$, $j_2$, $j_3$), $p^{out}_2=$($j_1$, $j_5$, $j_6$), $p^{out}_3=$($j_1$, $j_8$, $j_{9}$), $p^{out}_4=$($j_{1}$, $j_{11}$, $j_{12}$).
The outer angle represents the overall shape of the body part. $\theta_{in}$ and $\theta_{out}$ are simple; however, they effectively express scale-invariant human posture information.
 
{\bf{Local image patch feature:}}
To consider the local image where interactions actively occurred, the $(n {\times} n)$-size local image patch features are extracted from each joint of the potential major interaction area. This is motivated by the fact that most human interactions are performed using their hand or foot.
We chose the following five joints from each part, which are expected to have mainly contact when interaction occurs: $j_0$ (head position), $j_4$ (right hand), $j_7$ (left hand), $j_{10}$ (right foot), and $j_{13}$ (left foot). 
The input image patches are extracted where $Img(j,o)=([j_{i,x}-n/2:j_{i,x}+n/2], [j_{i,y}-n/2:j_{i,y}+n/2])$, when the center is in position $j_{i,x,y}$.
The feature representation for a specific joint $j_\alpha$ is the activation value of the last fully connected layer of the convolutional network $f_s$, with parameters $\Theta$, taking a local image patch $Img$ as the input, at the joint position $j_\alpha$ of object $o$: 
\begin{equation}
\label{LocalImgFeat}
\begin{aligned}
{\bf{pf}}_{j_\alpha} = f_s(Img({j_\alpha, o}); \Theta),\qquad \alpha=\{0, 4, 7, 10, 13\}.
\end{aligned}
\end{equation}
 
\subsection{Interacting body part attention}
In this section, we present the body joint based interacting body part attention mechanism.
The fundamental idea of interacting part-based attention is based on the assumption that the body parts composing each behavior will have different importance when human interaction occurs. 
For example, if two people shake hands, person \#1 will reach for person \#2 and person \#2 will similarly reach for person \#1. The hands of the two persons get closer, touch, and then become distant from each other.
However, if person \#1 punches person \#2, person \#1 reaches out to person \#2's head and person \#2 will be pushed back without motion toward person \#1. If person \#1 performs a push action, person \#2's response will look similar to a punch; however, person \#1 will reach out to person \#2's torso with two hands. These subtle differences between similar behaviors are maximized to generate discriminative features by assigning weight to interacting body parts.
 
We attempt to selectively focus on both the body part of the person who leads the interaction and the body part of the other person that is most involved. The different weights are multiplied by the terms of each part of each person. To express the movement of each part, the spatial difference for $t$ time segments is defined as follows:
 
\begin{equation}
\label{diffdistance}
\begin{aligned}
pw(p_i) = |d_{p,t}-d_{p,t-1}|,
\end{aligned}
\end{equation}
where $d$ denotes the relative distance between each pair of body parts of the interacting persons. The interacting body part attention between each person is calculated as follows:
 
\begin{equation}
\label{weight}
\begin{aligned}
\Lambda_{p,t}=S\times{\frac{\sum_{p}^{5}pw(p_i)}{pw(p_i)}}.
\end{aligned}
\end{equation}
The body part velocity $v_p^t$ and body part acceleration $\hat{v}_p^t$ are multiplied by the part weight $\Lambda_{p,t}$ to determine the weight of each part:
 
\begin{equation}
\label{weight_give}
\begin{aligned}
wv^t = \Lambda_{t}\otimes{v^t}, \\
w\hat{v}^t = \Lambda_{t}\otimes{\hat{v}^t}.
\end{aligned}
\end{equation}
A movement that is actively participating in the interaction, which is also called the interacting body part, is emphasized by representing individual human motion, using the weighted velocity of the body part $wv_p^t$ and the acceleration of the body part $w\hat{v}_p^t$.
 
The image patch feature vector from each joint is also multiplied by the weight. Since an interacting body part with high weight significant to the interaction, ${\bf{pf}}_{j_\alpha}^t$ also gives a high weight to the joint-based image feature extracted from the position of the body part as follows:
\begin{equation}
\label{weight}
{\bf{wf}}^t = \Lambda_t \otimes{\bf{pf}}^t.
\end{equation}
Each individual object's behavior with the interacting body part attention is represented by the combination of the following five variables of the five body parts: two posture variables $\theta_{in}$, $\theta_{out}$, two weighted motions variable $wv_p^t$, $w\hat{v}_j^t$, and one weighted local image feature ${\bf{wf}}^t$. 
 
\subsection{Full body feature stream}
Apart from the body joint-based feature and local image patch feature extraction, we also performed full-body image-based activity descriptor generation to consider the overall appearance change with co-occurrence of multi-person behaviors. Extracting a feature vector from a full-body image through CNN has proved its robustness in human activity recognition \cite{shu2017concurrence, lee2019prediction, ibrahim2016hierarchical, deng2016structure}. Because the occasional failure of body joint extraction may negatively affect the interaction representation, the imperfectness of local features is compensated by a full-body image descriptor. The full-body image descriptor generation method was inspired from the sub-volume co-occurrence matrix descriptor for human interaction prediction \cite{lee2019prediction}.
 
From the detected object region, we first extract the activation from the last fully connected layer of the inception-resnet-v2 network to obtain a feature vector of an object image. Subsequently, we generate a sub-volume for each object $\mathbf{f}$$_{oi}^{t} = [p, \delta{x}, \delta{y}]$, where $p$ denotes the average of the image representation feature vector in a sub-volume. A series of frame-level image feature vectors of object $oi$, at time $t$ for $l$ consecutive frames, are averaged into a single vector $\mathbf{f}_{oi}^t$.
Second, $K$-means clustering is performed on the training video to generate codewords $\{w_k\}_{k=1}^{K}$, where $k$ denotes the number of clusters. Each of the sub-volume features $\mathbf{f}_{oi}^{t}$ is assigned to the corresponding cluster $w_k$, following the BoW paradigm. The index of the corresponding cluster $k_{oi}^t$ is the codeword index, which is also the index of the row and column of the descriptor.
Finally, we construct a descriptor using sub-volume features. From each sub-volume of an object $v_{oi}^{t}= (\mathbf{f}, x, y, k)$, we first measure the spatial distance between sub-volumes $oi$ and $oj$ as follows:
 
\begin{equation}
dist_{oi,oj}^t = \sqrt{(x_{oi}^t-x_{oj}^t)^2+(y_{oi}^t-y_{oj}^t)^2}.
\end{equation}
The overall spatial distance between sub-volume $oi$ and the other $oj$ in segment $t$ for $\#pairs$, where $oj\neq oi$, is aggregated as follows:
 
\begin{equation}
r^t = \frac{1}{2}\sum_{oi}\sum_{oj\neq oi}dist_{oi,oj}^t.
\end{equation}
The difference in distance between sub-volume $oi$ and $oj$ to the global motion activation represents the participation ratio of the pair in segment $t$. The feature scoring function, based on sub-volume clustering, is calculated as follows:
 
\begin{equation}
f_p = log\left(\frac{||w_{oi}^t -\mathbf{f}_{oi}^t||+||w_{oj}^t-\mathbf{f}_{oj}^t||}{2} + \psi \right).
\end{equation}
After computing all required values between all sub-volumes, we finally construct the full-body image descriptor, as follows:
 
\begin{equation}
\label{SCMfinal}
M^t(k_{oi}^t,k_{oj}^t) =\frac{1}{N}\sum_{oi,oi\neq oj}\sum_{1:t} \frac{s_{oi}^t}{\epsilon^t}\frac{r^t}{dist_{oi,oj}^t}f_p(\mathbf{f}_{oi}^t, \mathbf{f}_{oj}^t),
\end{equation}
where $N$ is the normalization term. The value of the sub-volume between $oi$ and $oj$ is assigned to the full-body image descriptor using the corresponding cluster index, $k_{oi}^t$ and $k_{oj}^t$, of each sub-volume. Each of the descriptors is generated for every non-overlapped time step. Therefore, the descriptor is cumulatively constructed from the beginning.
 
\subsection{Feature fusion and classification}
In this section, we generate the combined vectors after the three-stream feature generation. 
From an individual object, we extract five local image features ${\bf{wf}}^t$, which represent the local image appearance. We select a feature vector ${\bf{wf}}^t_{mp}$ from the most activate body part, where the $\Lambda_p$ is maximized. The local image patch feature is concatenated with the motion feature and posture feature, as a combined vector of individual $\#m$, ${\bf{cf}}_m = [{\bf{wf}}_{mp}, \theta_{in}, \theta_{out}, wv_p, w\hat{v}_j]$.
The representation of individual $\#m$ was the fed into the sub-LSTM $\#m$ to model the temporal dynamics of each person.
At each time step, the hidden unit of the sub-LSTM $\#m$, $h^t_m$ is concatenated with a full body image co-occurrence descriptor $M^t$ as ${\bf{ff}}^t = [h^t_1, h^t_2, ... ,h^t_m, M^t]$.
Finally, the concatenated vector ${\bf{ff}}^t$ is used as the input to the LSTM and the activity classification task is the output of the LSTM after $t$ segments.
 
\section{Experiments}
\label{Experiment}
\subsection{Datasets}
We validated the effectiveness of the proposed method by comparing with the state-of-the-art methods on the (1) BIT-Interaction dataset \cite{kong2012learning}; (2) UT-Interaction dataset \cite{UT-Interaction-Data}; (3) VIRAT 1.0 Ground dataset and VIRAT 2.0 Ground dataset \cite{oh2011large}.
The first two datasets are common and widely used in human-human interaction recognition research. 
The VIRAT 1.0 and VIRAT 2.0 Ground dataset consists of human-vehicle interaction tasks, which we chose with the aim of exhibiting the extensibility of our work.
 
\begin{figure}
\centering
\subfigure[BIT: Bow]{
\includegraphics[width=.35\linewidth]{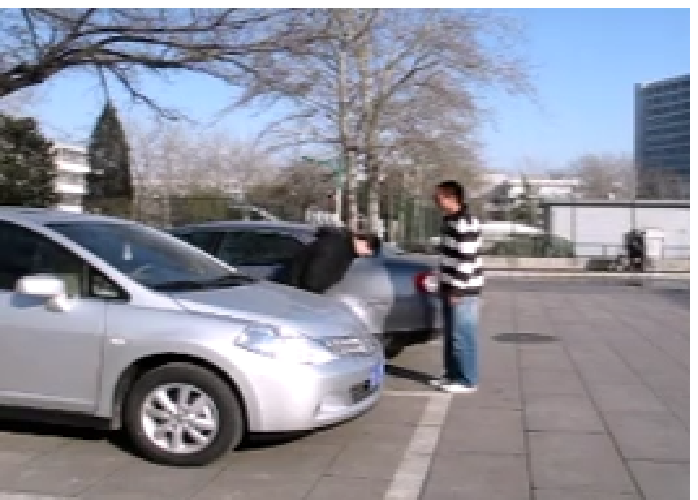}\label{bit1}}
\subfigure[BIT: Boxing]{
\includegraphics[width=.35\linewidth]{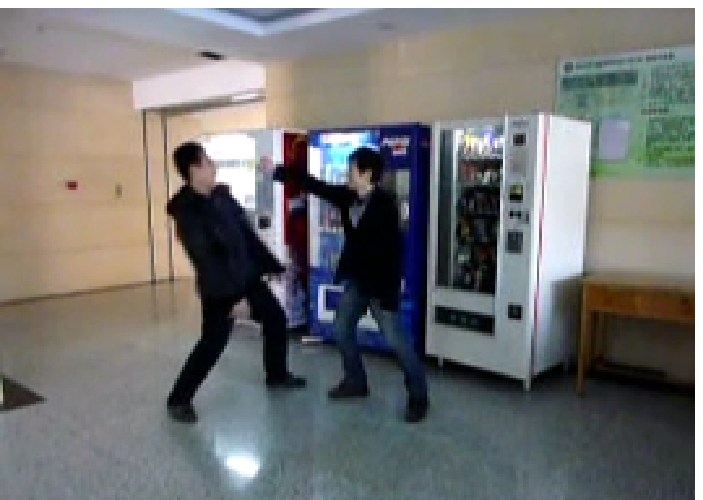}\label{bit2}}
\subfigure[UT-set \#1: Push]{
\includegraphics[width=.35\linewidth]{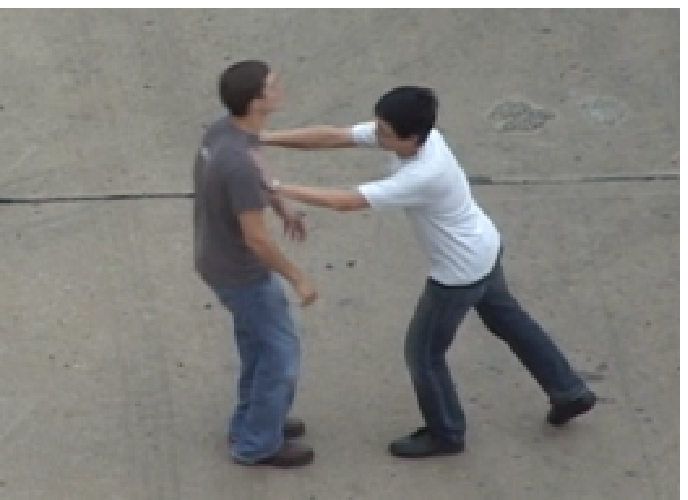}\label{ut11}}
\subfigure[UT-set \#1: Kick]{
\includegraphics[width=.35\linewidth]{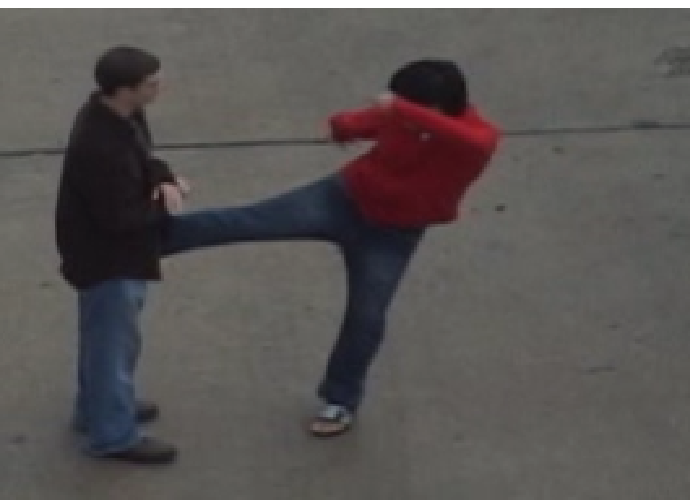}\label{ut12}}
\subfigure[UT-set \#2: Push]{
\includegraphics[width=.35\linewidth]{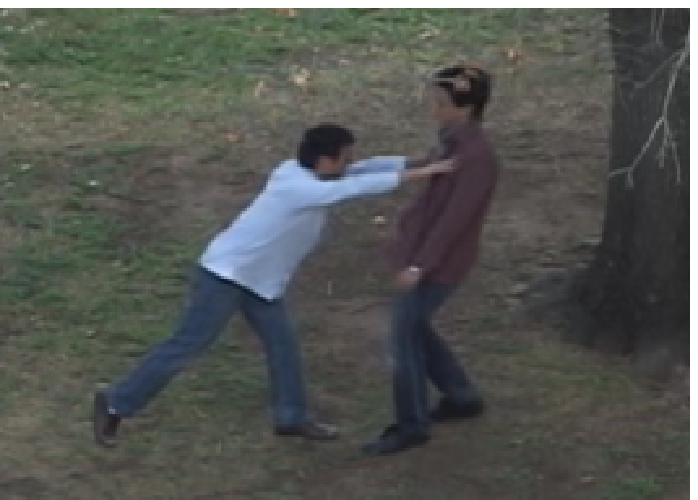}\label{ut21}}
\subfigure[UT-set \#2: Kick]{
\includegraphics[width=.35\linewidth]{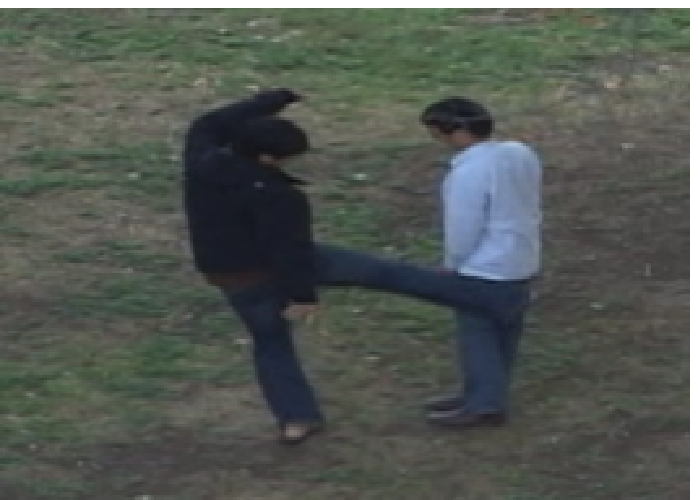}\label{ut22}}
\subfigure[VIRAT 1.0: GOV]{
\includegraphics[width=.35\linewidth]{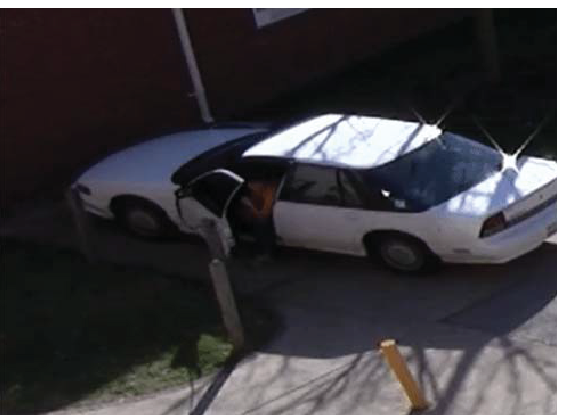}\label{virat10}}
\subfigure[VIRAT 2.0: OAT]{
\includegraphics[width=.35\linewidth]{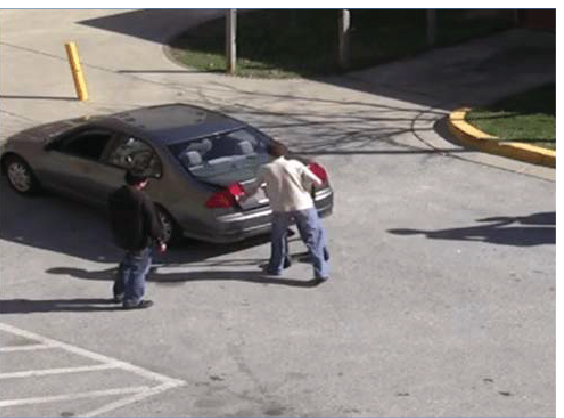}\label{virat20}}
\caption{Sample frames of the BIT-Interaction dataset (a)-(b), UT-Interaction dataset set-\#1 (c)-(d), set-\#2 (e-f), VIRAT 1.0 (g), and VIRAT 2.0 (h) Ground dataset.}
\label{bit-sample}
\end{figure}
 
{\bf{The BIT-Interaction dataset}} used in the experimental evaluation consists of eight classes of human interactions: bow, boxing, handshake, high-five, hug, kick, pat, and push. Each class contains 50 clips. The videos were captured in a very realistic environment, including partial occlusion, movement, complex background, varying sizes, view-point changes, and lighting changes.
For the evaluation, the training set is composed of 34 videos per class (272 in total), and the remaining 16 videos per class (128 in total) are used as the test set following in \cite{kong2014interactive, kong2016max, kong2015close, lee2019prediction, shu2017concurrence}. Fig. \ref{bit1}-\ref{bit2} shows an example snapshot of the BIT-Interaction dataset. In \ref{bit1}, the lower half of the body of the person bowing on the left is occluded by a parked vehicle. There is a vending machine in the background in \ref{bit2}. Both images show the environmental difficulties in analyzing the human activity.
 
{\bf{The UT-Interaction dataset}} used in the experimental evaluation consists of six classes of human interactions: push, kick, hug, point, punch, and handshake. The dataset is composed of two sets of videos; UT-set \#1 and UT-set \#2, which were captured in different environments. Each class contains 10 clips for each set. We performed a leave-one-out cross-validation for the evaluation of each set, following competing methods \cite{ryoo2011human, raptis2013poselet, kong2015close, kong2014interactive, el2020learning, shu2017concurrence, lee2019prediction}. The set \#1 videos were captured in a parking lot background (Fig. \ref{ut11}-\ref{ut12}). However, the backgrounds in set \#2 of the UT-Interaction dataset (Fig. \ref{ut21}-\ref{ut22}) consisted of grass and shaking twigs, which may add noise to local patches of interest points.
 
{\bf{VIRAT 1.0 and VIRAT 2.0 Ground datasets}} used in the experimental evaluation, consist of six classes of human-object interactions: a person opening a vehicle trunk (OAT), a person closing a vehicle trunk (CAT), a person loading an object to a vehicle (LAV), a person unloading an object from a vehicle (UAV), a person getting into a vehicle (GIV), and a person getting out of a vehicle (GOV). The VIRAT 1.0 Ground dataset and VIRAT 2.0 Ground dataset consist of approximately 3 hours and 8 hours of video in several parking lot backgrounds, respectively. 
Fig. \ref{virat10} demonstrates that the acting person is not fully observable until he gets out of the vehicle in the shadow.
The carrying object is fully occluded by the acting person in \ref{virat20}, which can disturb the understanding of the context of opening a vehicle trunk.
For the evaluation, half of the video clips were used as the training set, and the remaining half were used as a testing set following previous studies \cite{zhu2013context, wang2014hierarchical, wang2016hierarchical, lee2019prediction, ajmal2019recognizing}.
 
\subsection{Implementation details}
In this experiment, the initial pose estimation of human objects over all video frames was done using OpenPose \cite{cao2018openpose}. The bounding boxes of person are detected by Faster-RCNN \cite{ren2015faster} with the Inception-resent-v2 network \cite{szegedy2017inception}.
To extract image patch features from joints and full body image features, we used the weight of the pre-trained Inception-resnet-v2 network, as a backbone network.
The learning rate was set to $0.5\times10^{-4}$ and the lambda\_loss set to $0.5\times10^{-3}$.
The length of the time step was set to 20 and the minibatch of size 32. 
We set the $K$ values of the full-body feature clustering is set to 20.
The number of memory cell nodes of the sub-LSTMs and the second LSTM was set to 200 and 625, respectively. The weights and biases of the LSTMs are initialized with random values having normal distributions. 
We used Adam optimizer \cite{kingma2014adam} for all experiments.
We implemented all modules in this framework using Tensorflow \cite{tensorflow2015-whitepaper}, and an NVIDIA Titan XP was used to run the experiments.

In these experiments, we additionally conduct two baseline methods to show the effect of each feature stream on the proposed framework as follows:
 
{\bf{baseline \#1:}} This baseline only used body joint-based features, with attention from the joint-based feature stream and local image patch feature stream. The interacting body part attention is applied to this baseline. The concatenated features at all time steps are only pooled from the LSTM of an individual object.
 
{\bf{baseline \#2:}} This baseline used the full-body feature, without the body joint-based feature and interacting body part attention. The co-occurrence descriptor of the image-based full-body feature stream was directly fed into final LSTM. This baseline is similar to that of the sub-volume co-occurrence matrix \cite{lee2019prediction}.
 
\subsection{Results on BIT-Interaction dataset}
The experimental results on the BIT-Interaction dataset are compared with the latest studies, i.e. dynamic BoW \cite{ryoo2011human}, mixture of training video segment sparse coding (MSSC) \cite{cao2013recognize}, multiple temporal scale support vector machine (MTSSVM) \cite{kong2014discriminative}, max-margin action prediction machine (MMAPM) \cite{kong2016max}, Lan \etal \cite{lan2011discriminative}, Liu \etal \cite{liu2011recognizing}, Kong \etal \cite{kong2014interactive, kong2015close}, Donahue \etal \cite{donahue2015long}, sub-volume co-occurrence matrix (SCM) \cite{lee2019prediction}, concurrence-aware long short-term sub-memories (Co-LSTM) \cite{shu2017concurrence}, and hierarchical long short-term concurrent memory(H-LSTCM) \cite{shu2019hierarchical}. We also compared the proposed method with two baseline performances.
 
\begin{table}[h]
\begin{center}
\caption{Comparison of the recognition results on the BIT-Interaction dataset}
\label{tableBIT}
\begin{tabular}{lcccc}
\hline
Method & Accuracy(\%) \\
\hline
Dynamic BoW \cite{ryoo2011human} & 53.13\\
MSSC \cite{cao2013recognize} & 67.97\\
MTSSVM \cite{kong2014discriminative} & 76.56 \\
MMAPM \cite{kong2016max} & 79.69\\
Lan \etal \cite{lan2011discriminative} & 82.03\\
Liu \etal \cite{liu2011recognizing} & 84.37\\
Kong \etal \cite{kong2015close} & 85.38\\
Kong \etal \cite{kong2014interactive} & 90.63\\
Donahue \etal \cite{donahue2015long} & 80.13\\
SCM \cite{lee2019prediction} & 88.70\\
Co-LSTM \cite{shu2017concurrence} & 92.88\\
H-LSTCM \cite{shu2019hierarchical} & 94.03 \\
\hline
Baseline \#1 & 77.25\\
Baseline \#2 & 88.50\\
Proposed Method & 93.10\\
\hline
\end{tabular}
\end{center}
\end{table}

Table \ref{tableBIT} shows the classification results of the proposed method for a quantitative comparison with competing methods.
The proposed method achieved higher recognition accuracy on average than all baseline methods and most of the competing approaches. In this experiment, the very recent H-LSTCM \cite{shu2019hierarchical} shows 0.8\% higher accuracy owing to the dynamic inter-related representation of the Co-LSTM unit, but the proposed method also achieved a comparable accuracy with remarkable representation power itself. 
Specifically, the proposed method with all three feature streams achieved a higher accuracy than baseline \#1 and baseline \#2, by combining implicit and explicit representation. Thus, the body joint-based representation with attention complements the lack of high-level information on the low-level image feature representation.It is better to use the high-level information than to utilize the low-level image features alone. The comprehensive consideration of individual human motions, co-occurrence of overall appearance, and local importance representation let achieve robust human interaction recognition.
 
\subsubsection{Ablation study of body joint extraction} 
We also conducted an experiment to show the effectiveness of proposed indexes of body joints with five body parts. In this experiment, we extracted motion features only using $j^{ori}_i$, and local image patches were extracted at the same coordinates. If the pose estimation was failed at a certain frame, the coordinate of joints at the last successful frame was used. 
We performed an experiment in four different settings; original body joints/the proposed indexes of body joints with/without the interacting body part attention.
In the ablation study on the BIT-Interaction dataset, the original body joints from pose estimation achieved 67.25\% without interacting body part attention and 74.37\% with the interacting body part attention, respectively. Meanwhile, the proposed indexes of body joints achieved 70.67\% and 77.25\% accuracy at the same condition. The stable extraction of each body joint also influences the weight of interacting body part attention. The result shows that the proposed index with five body parts shows slightly better performance than the original indexes, and the proposed interacting body part attention also effectively works for a fine-grained representation of the interaction.

\subsection{Results on UT-Interaction dataset}
{\bf{UT-set \#1:}} We compared the recognition accuracy of the proposed method with a competing state-of-the-art method, including some traditional methods and two baselines, i.e. dynamic BoW \cite{ryoo2011human}, Lan \etal \cite{lan2014hierarchical}, MSSC \cite{cao2013recognize}, MMAPM \cite{kong2016max}, Wang \etal \cite{wang2016hierarchical}, Donahue \etal \cite{donahue2015long}, SCM \cite{lee2019prediction}, Co-LSTM \cite{shu2017concurrence}, H-LSTCM \cite{shu2019hierarchical}, Mahmood \etal \cite{mahmood2018robust}, and Slimani \etal \cite{el2020learning}.
 
\begin{table}[h]
\begin{center}
\caption{Comparison of the recognition results on the UT-Interaction dataset (set \#1).}
\label{UTSet1}
\begin{tabular}{lcccc}
\hline
Method & Accuracy(\%) \\
\hline
Bag-of-Words(BoW) & 81.67\\
Integral BoW \cite{ryoo2011human} & 81.70\\
Dynamic BoW \cite{ryoo2011human} & 85.00\\
MSSC \cite{cao2013recognize} & 83.33 \\
MMAPM \cite{kong2016max} & 95.00 \\
Poselet \cite{raptis2013poselet} & 93.30 \\
Kong \etal \cite{kong2015close} & 93.33\\
Kong \etal \cite{kong2014interactive} & 91.67\\
Donahue \etal \cite{donahue2015long} & 85.00 \\
SCM \etal \cite{lee2019prediction} & 90.22 \\
Co-LSTM \cite{shu2017concurrence} & 95.00 \\
Wang \etal \cite{wang2016hierarchical} & 95.00 \\
H-LSTCM \cite{shu2019hierarchical} & 98.33 \\
Mahmood \etal \cite{mahmood2018robust} & 83.50 \\
Slimani \etal \cite{el2020learning} & 90.00 \\
\hline
Baseline \#1 & 88.50\\
Baseline \#2 & 90.00\\
Proposed Method & 93.33\\
\hline
\end{tabular}
\end{center}
\end{table}

The comparison of the interaction recognition performance between the proposed method with two baselines and competing methods, on the UT-set \#1, is shown in Table \ref{UTSet1}. 
It can be seen that the combination of interacting body part attention and co-occurrence descriptor could improve the performance, which prove that specific information from different types of features can complement each other.
In this experiment, the proposed method achieved 93.22\% recognition accuracy, which is slightly lower than the competing state-of-the-art methods \cite{wang2014hierarchical,kong2016max,shu2017concurrence, shu2019hierarchical}. 
We believe that the reason for very high performances in UT-set \#1 is the superiority of the competing methods as well as the environmental simplicity. The UT-set \#1 is recorded against a gray asphalt background, which allows for the clear identification of the appearance of a person, without noise. Here, it is noted that Wang \etal \cite{wang2016hierarchical} adopt deep context feature on the event neighborhood, the size of which requires manual definition. Thus, all the competing methods achieved high performance, with ideal and rich appearance information, while the proposed method experienced difficulty in extracting body joints from some frames, owing to their occlusion.

{\bf{UT-set \#2:}}
We separately compared the recognition accuracy of the proposed method on the UT-set \#2, for a fair comparison. Table \ref{UTSet2} shows the performance of competing state-of-the-art methods with some traditional methods and two baselines, i.e., dynamic BoW \cite{ryoo2011human}, Lan \etal \cite{lan2014hierarchical}, MSSC \cite{cao2013recognize}, SC \cite{cao2013recognize}, MMAPM \cite{kong2016max}, SCM \cite{lee2019prediction}, Mahmood \etal \cite{mahmood2018robust}, and Slimani \etal \cite{el2020learning}.
 
\begin{table}[h]
\begin{center}
\caption{Comparison of the recognition results on the UT-Interaction dataset (set \#2).}
\label{UTSet2}
\begin{tabular}{lcccc}
\hline
Method & Accuracy(\%) \\
\hline
Bag-of-Words(BoW) & 80.00\\
Dynamic BoW \cite{ryoo2011human} & 70.00\\
Lan \etal \cite{lan2014hierarchical}& 83.33\\
MSSC \cite{cao2013recognize} & 81.67\\
SC \cite{cao2013recognize} & 80.00\\
MMAPM \cite{kong2016max} & 86.67\\
SCM \cite{lee2019prediction} & 89.40\\
Mahmood \etal \cite{mahmood2018robust} & 72.50\\
Slimani \etal \cite{el2020learning} & 83.90 \\
\hline
Baseline \#1 & 78.33\\
Baseline \#2 & 88.50\\
Proposed Method & 91.33\\
\hline
\end{tabular}
\end{center}
\end{table}

We can see that different from UT-set \#1, the proposed method outperformed all the competing methods with 91.33\% recognition accuracy on UT-set \#2.
The proposed method exhibits outstanding recognition performance considering that the accuracy of the competing methods on UT-set \#2 were decreased approximately 2\%-15\% on average compare to UT-set \#1. 
The recognition performances of dynamic BoW \cite{ryoo2011human}, MSSC \cite{cao2013recognize}, MMAPM \cite{kong2016max}, Mahmood \etal \cite{mahmood2018robust}, Slimani \etal \cite{el2020learning} decreased from 85.00\%, 83.33\%, 95.00\%, 83.50\%, and 90.00\% to 70.00\%, 81.67\%, 86.67\%, 72.50\%, and 83.90\%, respectively.
This is because UT-set \#2 has more complicated background than UT-set \#1; thus, the proposed method of using human structural features through joint estimation performs better than the competing methods. We believe that the proposed method is highly effective, considering the complexities of the environmental change in real-world scenarios.

\subsection{Results on VIRAT Ground datasets}
We also evaluate the proposed framework on the VIRAT 1.0 Ground dataset and VIRAT 2.0 Ground dataset, to show the extensibility of our method. We compared the performance with Zhu \etal \cite{zhu2013context}, Bayesian network (BN) \cite{wang2014hierarchical}, a deep hierarchical context model (DHCM) \cite{wang2016hierarchical}, SCM \cite{lee2019prediction}, and WSCF \cite{ajmal2019recognizing}.
Because these two datasets consist of human-object interaction tasks, each human and object (mainly vehicle or box) are treated as separate objects. Here, we should note that in this experiment, the object does not have a body joint; the center of the detected object is set to the coordinates of the interacting joint. In the body joint extraction procedure, the failure of the head and torso estimation of non-human objects was exceptionally handled in this experiment.
 
\begin{table}[h]
\begin{center}
\caption{Comparison of the activity recognition results on the VIRAT 1.0 and VIRAT 2.0 Ground dataset.}
\label{VIRAT10}
\begin{tabular}{lcccccccc}
\hline 
Method & VIRAT 1.0(\%) & VIRAT 2.0(\%)\\
\hline
SVM-STIP & - & 41.74 \\
Zhu \etal \cite{zhu2013context} & 62.90 & - \\
BN \cite{wang2014hierarchical} &65.80 & 58.70 \\
DHCM \cite{wang2016hierarchical} & 69.90 & 66.45 \\ 
SCM \cite{lee2019prediction} & 76.30 & 79.53 \\
WSCF \cite{ajmal2019recognizing} & 73.14 & 74.41 \\
\hline
Baseline \#1 & 65.58 & 61.83 \\
Baseline \#2 & 75.33 & 77.83 \\
Proposed Method & 79.83 & 81.50 \\
\hline
\end{tabular}
\end{center}
\end{table}
 
Table \ref{VIRAT10} lists the recognition accuracies of human-vehicle interactions, using VIRAT 1.0 Ground dataset and VIRAT 2.0 Ground dataset. The proposed method outperformed all the competing methods, with recognition accuraccy of 79.83\% and 81.50\%, respectively. 
 
\begin{figure}
\centering
\includegraphics[width=1.0\linewidth]{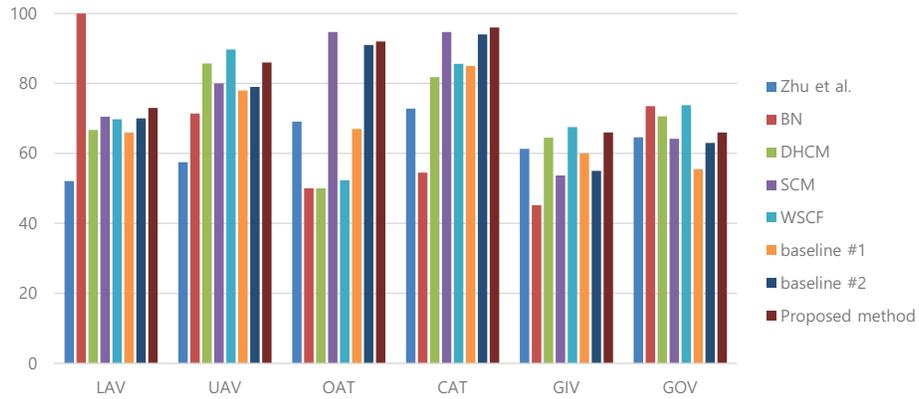}
\caption{Comparison of proposed method with baselines and competing method on the VIRAT 1.0 Ground dataset for each class}
\label{exsub1}
\end{figure}
 
\begin{figure}
\centering
\includegraphics[width=1.0\linewidth]{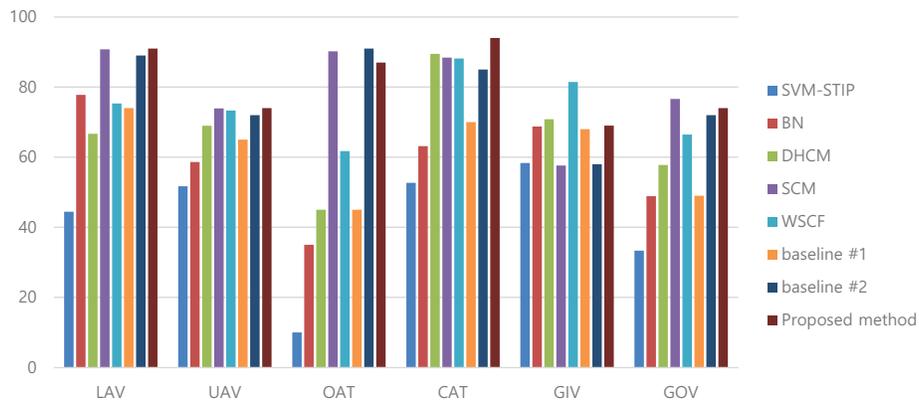}
\caption{Comparison of proposed method with baselines and competing method on the VIRAT 2.0 Ground dataset for each class}
\label{exsub2}
\end{figure}
 
In both datasets, the baseline \#1 shows a lower performance. Owing to the characteristics of the type of interaction in the dataset, body joint extraction failures occur very frequently when occlusion occurs with the vehicle. Moreover, the vehicle has no reaction to human action at the joints. The absence of mutual interacting body parts also makes it difficult to use a local image patch and body joint-based feature.
However, the proposed method shows a much higher performance than baselines \#1 and \#2, considering the full-body image together. The representation power of the full-body image feature is sufficient to describe the behavior of the acting objects. 
Moreover, the local image patch feature, extracted from the joint of the acting object, provides information about the directly interacting part of the vehicle, such as the trunk or the door. We can observe the recognition results of the VIRAT 1.0 and VIRAT 2.0 Ground dataset for each class in Fig. \ref{exsub1} and Fig. \ref{exsub2}, respectively. We can see that the baseline \#1 result does not show the worst accuracies on some interactions, despite the difficulty in body joint extraction.
We conduct a similar experiment on the VIRAT 2.0 Ground dataset. As expected, the proposed method achieved a better performance than other methods. Even though, for the same reasons mentioned for the VIRAT 1.0 Ground dataset, baseline \#1 shows a lower accuracy, the propose framework achieved the best performance by complementary modeling of human activity with full-body co-occurrence with interacting body part attention.
 
\section{Conclusion and Future Work}
\label{conclusion}
In this paper, we present a novel framework for human interaction recognition, which considers both the implicit and explicit representations of the human behavior.
Specifically, the subtle differences in behaviors of interacting persons are captured through interacting body part attention, which maximizes the discriminative power of feature representation.
The complex activity of two or more people interacting with each other requires a higher-level understanding of both individual human behavior and their relationships.
Both implicit and explicit representations are successfully learned through the individual behavior of the person represented, based on interacting body part attention with their local image patch, after stabilized body joint extraction.
The image-based full-body feature models the interaction, based on the co-occurrence of human behaviors and their activations.
Each of the three features feeds into the LSTM to model individual level behavior and video level interaction, while working complementarily.
Experimental results on four publicly accessible datasets, i.e., the BIT-Interaction dataset, UT-Interaction dataset, VIRAT 1.0 Ground dataset, and the VIRAT 2.0 Ground dataset, demonstrated the superiority of the proposed method compared to state-of-the-art methods.
The attention to the interacting body part showed a weakness when the human body joint estimation fails owing to heavy occlusion.
Nevertheless, if the body joint is extracted even a little including neighbor frames, it demonstrates good results, owing to the local image patch feature with actively interacting body parts.
The proposed method provides a convenient and straightforward understanding of accurate human interaction recognition. However, the use of the proposed method is not established for more than three-person interaction such as collective activity, currently. We expect that the multi-person interaction problem can be addressed by adaptive fusing the result of individual human action modeling. Thus, the extension of the proposed method to the complex multi-person activity will be our future work.

\section*{Acknowledgment}
This work was supported by Institute for Information \& Communications Technology Planning \& Evaluation(IITP) grant funded by the Korea government (MSIT) [No. 2019-0-00079, Department of Artificial Intelligence, Korea University] and [No. 2014-0-00059, Development of Predictive Visual Intelligence Technology].

\bibliography{mybibfile.bbl}
 
\end{document}